\documentclass[10pt,twocolumn,letterpaper]{article}

\usepackage{cvpr}              %

\definecolor{cvprblue}{rgb}{0.21,0.49,0.74}
\usepackage[pagebackref,breaklinks,colorlinks,allcolors=cvprblue]{hyperref}
\usepackage{bm}
\usepackage{amsmath}
\usepackage{amssymb}

\usepackage{lineno}

\usepackage{makecell}
\usepackage{arydshln}
\usepackage[accsupp]{axessibility}  %

\newcommand{\cmdSOL}{\langle\mathtt{SOL}\rangle}
\newcommand{\cmdLine}{\mathtt{L}}
\newcommand{\cmdArc}{\mathtt{A}}
\newcommand{\cmdCirc}{\mathtt{R}}
\newcommand{\cmdExt}{\mathtt{E}}
\newcommand{\cmdEOS}{\langle\mathtt{EOS}\rangle}
\newcommand{\cmdNum}{N_c}

\newcommand{\embDim}{d_\txt{E}}
\newcommand{\embCmd}{\bm{e}_i^{\text{cmd}}}
\newcommand{\embPm}{\bm{e}_i^{\text{param}}}
\newcommand{\embPos}{\bm{e}_i^{\text{pos}}}

\newcommand{\txt}[1]{\textrm{#1}}

\newcommand{\z}{\ensuremath{\boldsymbol{z}}}
\newcommand{\zw}{\ensuremath{\boldsymbol{z}^w}}
\newcommand{\zl}{\ensuremath{\boldsymbol{z}^l}}

\newcommand{\calvar}[1]{\ensuremath{\mathcal{#1}}}
\newcommand{\calD}{\calvar{D}}

\newcommand{\noise}{\ensuremath{\boldsymbol{\epsilon}}}

\newcommand\blfootnote[1]{%
  \begingroup
  \renewcommand\thefootnote{}\footnote{#1}%
  \addtocounter{footnote}{-1}%
  \endgroup
}

\def\paperID{12861} %
\def\confName{CVPR}
\def\confYear{2025}

\title{CADCrafter: Generating Computer-Aided Design Models from Unconstrained Images}

\author{Cheng Chen$^{1,2}$\textsuperscript{*}, Jiacheng Wei$^{1}$\textsuperscript{*}, Tianrun Chen$^{3,7}$\textsuperscript{\dag}, Chi Zhang$^{5}$,  Xiaofeng Yang$^{1}$, Shangzhan Zhang$^{7}$,\\ Bingchen Yang$^{1}$, Chuan-Sheng Foo$^{2,6}$, Guosheng Lin$^{1}$, Qixing Huang$^{4}$, Fayao Liu$^{2}$\textsuperscript{\dag} \\
$^{1}$Nanyang Technological University,  $^{2}$Institute for Infocomm Research, A*STAR, Singapore \\ 
  $^{3}$ KOKONI3D, Moxin (Huzhou) Technology Co., LTD., $^{4}$ The University of Texas at Austin \\  $^{5}$Westlake University,  $^{6}$Centre for Frontier AI Research, A*STAR, Singapore\\ $^{7}$ Zhejiang University \\
 \tt\small {\{cheng021, jiacheng.wei\}@ntu.edu.sg, tianrun.chen@kokoni3d.com, {fayaoliu}@gmail.com}}

\begin{document}
\maketitle
\begin{abstract}
\blfootnote{$^*$ The first two authors contributed equally to this work.}
\blfootnote{$^{\dag}$ Corresponding authors.}

Creating CAD digital twins from the physical world is crucial for manufacturing, design, and simulation. However, current methods typically rely on costly 3D scanning with labor-intensive post-processing. To provide a user-friendly design process, we explore the problem of reverse engineering from unconstrained real-world CAD images that can be easily captured by users of all experiences. However, the scarcity of real-world CAD data poses challenges in directly training such models. To tackle these challenges, we propose CADCrafter, an image-to-parametric CAD model generation framework that trains solely on synthetic textureless CAD data while testing on real-world images. To bridge the significant representation disparity between images and parametric CAD models, we introduce a geometry encoder to accurately capture diverse geometric features. Moreover, the texture-invariant properties of the geometric features can also facilitate the generalization to real-world scenarios. Since compiling CAD parameter sequences into explicit CAD models is a non-differentiable process, the network training inherently lacks explicit geometric supervision. To impose geometric validity constraints, we employ direct preference optimization (DPO) to fine-tune our model with the automatic code checker feedback on CAD sequence quality. Furthermore, we collected a real-world dataset, comprised of multi-view images and corresponding CAD command sequence pairs, to evaluate our method. Experimental results demonstrate that our approach can robustly handle real unconstrained CAD images, and even generalize to unseen general objects.

\vspace{-10pt}
\end{abstract}

\section{Introduction} 
Computer-aided design (CAD) provides fundamental mechanical components that are essential to create shapes and mechanisms in all manufacturing and design applications. Parametric CAD command sequences enable precise control over shapes and facilitate effortless future modifications on size and scales. However, manual creation of CAD command sequences is tedious and time-consuming, leading to reverse engineering studies to recover CAD design procedures from existing CAD models. Current research focuses predominantly on reconstructing CAD command sequences from 3D representations such as B-Reps~\cite{willis2021engineering, xu2021inferring}, point clouds~\cite{ma2024draw, wu2021deepcad}, and voxels~\cite{li2023secad, li2024sfmcad}. These forms are typically derived from synthetic digital data or from high-quality 3D reconstructions obtained using costly 3D sensors. This dependency on sophisticated data and expensive technology limits the feasibility of these methods in practical everyday applications.

\begin{figure}[t]
  \centering
   \includegraphics[width=1.0\linewidth]{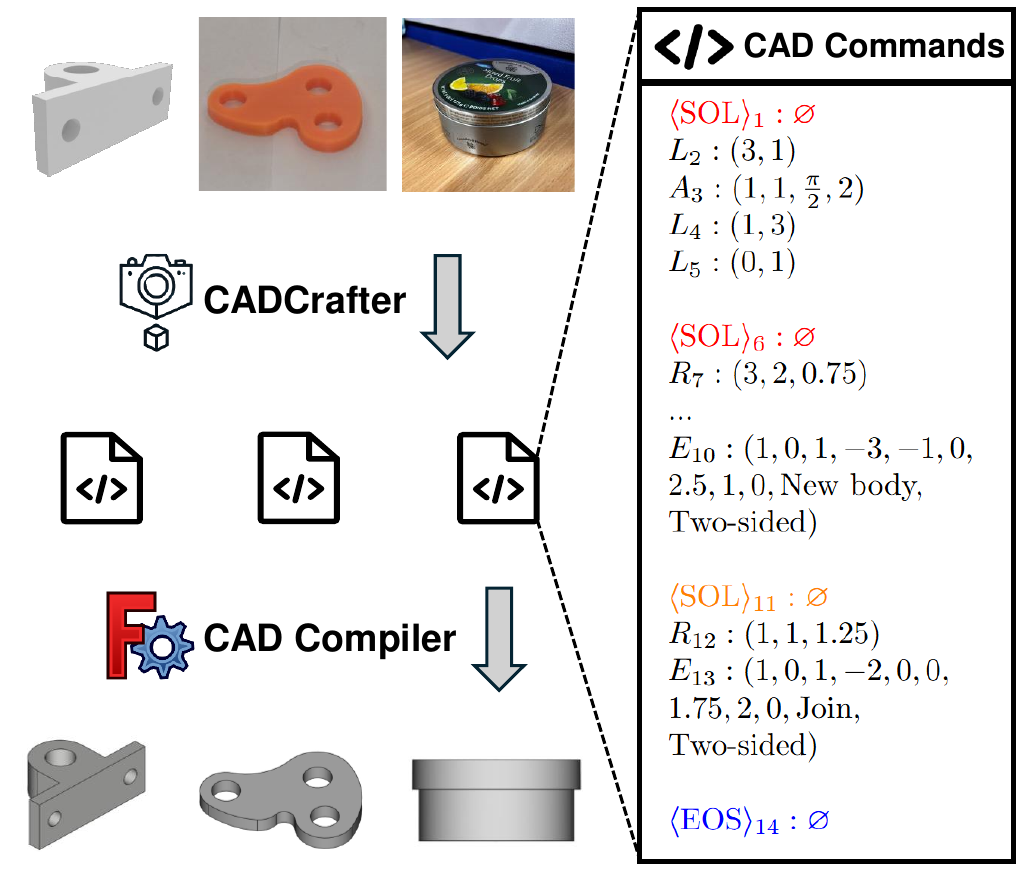}
\vspace{-10pt}
   \caption{Our proposed CADCrafter can generate CAD command sequences from unconstrained multi-domain images, including (from left to right) synthetic data renderings, 3D-printed CAD models, and unseen general objects. These generated CAD commands can then be compiled into 3D CAD models. Notably, our model is trained solely on synthetic data renderings.}
   \label{fig:teaser}
   \vspace{-20pt}
\end{figure}

Recent advances in generative models have facilitated the generation and reconstruction of images to 3D~\cite{liu2024one,liu2024meshformer, long2024wonder3d,hong2023lrm} with ease. However, these methods often yield 3D shapes with rough surfaces and indistinct, blurred edges, failing to accurately replicate geometric standards such as rectangles or circles. Moreover, the 3D shapes produced by these methods are difficult to edit and lack the precision required for direct use in manufacturing. %

This prompts us to investigate the feasibility of directly generating editable CAD command sequences from images, as illustrated in Figure~\ref{fig:teaser}. However, the task is particularly difficult due to the significant representational and domain gap between the two modalities, where CAD commands consist of a mix of discrete geometric operations and continuous parameters while images capture raw appearance with limited spatial information. The challenge is amplified with in-the-wild, unconstrained images. These images frequently exhibit variability in camera poses, lighting conditions, and noise, as well as various materials and textures of the objects depicted. Another challenge in developing this system lies in the difficulty of collecting paired CAD commands and image data from real-world scenarios, making it necessary to rely on synthetic datasets for training. However, models trained solely on synthetic data often underperform with real-world data. Therefore, there is a critical need for a method to bridge this gap, ensuring that approaches trained on synthetic data can perform effectively on both synthetic and real-world unconstrained images.

\begin{figure*}[t]
  \centering
   \includegraphics[width=\linewidth]{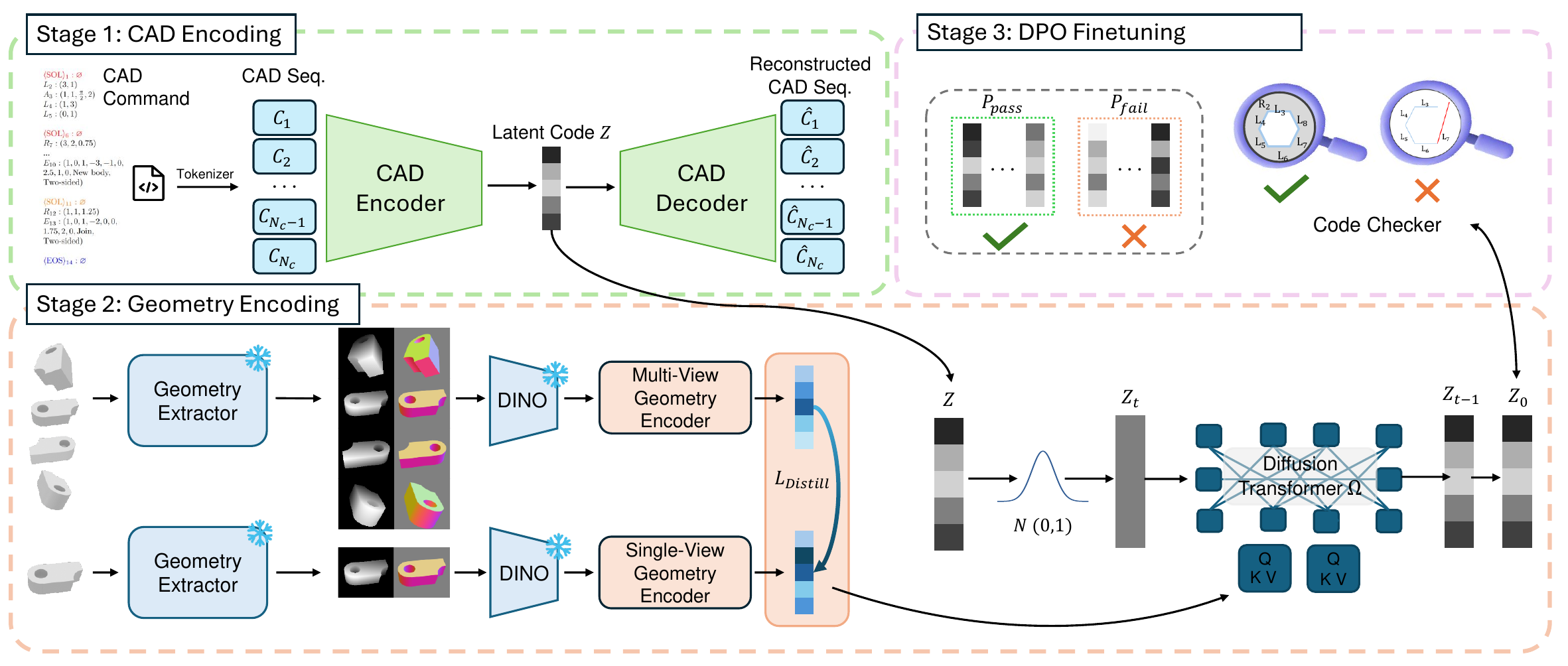}

 \vspace{-10pt}
   \caption{The training pipeline comprises three stages. In the first, a transformer autoencoder reconstructs CAD command sequences into a latent space. Second, we extract depth and normal using a pre-trained geometric extractor, the encoded features serve as conditions in the latent diffusion model; the multi-view geometric encoders and the latent diffusion model are jointly trained. Later, a single-view geometry encoder is trained by distilling knowledge from the multi-view encoder to enhance robustness. Third, we develop a geometry validity-based code checker and fine-tune the diffusion model with direct preference optimization (DPO) to improve generation quality and accuracy.}

 \vspace{-10pt}
   \label{fig:pipeline}
\end{figure*}

To tackle these challenges, we propose CADCrafter, a method engineered to directly generate CAD command sequences from both multi-view and single-view unconstrained images. Specifically, as shown in Figure \ref{fig:pipeline}, we first train a transformer-based autoencoder to map the CAD tokens to a latent space and then reconstruct them. Then, we apply a latent diffusion transformer to denoise the latent CAD codes conditioned on the input images.
Unlike traditional latent diffusion architectures \cite{ye2023ip} that rely directly on image features as conditions, our approach utilizes geometric features of images, specifically depth and normal maps. There are two main benefits of geometric features: they enhance geometry representations to boost accurate command prediction and are invariant to the textural gap between synthetic data and real-world images. 
Additionally, different modalities of geometry features capture various perspectives of an object, with each modal providing unique geometric information. To capitalize on this, we designed a geometry encoder that adaptively consolidates geometric data from each modality.

When generating CAD commands, the non-differentiable nature of the CAD compiler makes it inherently challenging to directly incorporate geometric constraints. Due to the geometry precision required in CAD models, inaccurate commands may fail to compile into a valid CAD model,  as shown in Figure \ref{fig:code_checker}.

To implicitly learn the correct shape pattern of CAD models,  we enhance the latent diffusion model with additional constraints and regularization, c.f.~\cite{DBLP:conf/cvpr/DongZ0YZDBH24}. Drawing inspiration from reinforcement learning with human feedback (RLHF) \cite{bai2022training}, we implement a code checker to improve the validity of denoised latent codes. Specifically, we deploy the CAD compiler as an automatic checker to categorize codes as valid or invalid. Subsequently, we fine-tune the diffusion model using these categorized sets through direct preference optimization \cite{rafailov2024direct, wallace2024diffusion} to improve the generation quality and accuracy. 

Since single-view images inherently lack complete information about the unseen part of a 3D object, instead of training separate models for multi-view and single-view inputs, we distill the comprehensive knowledge from our pre-trained multi-view geometry encoder into a single-view geometric encoder by aligning their feature representations. This enables the model to learn the mapping from single-view to multi-view input, thereby enhancing both the accuracy and robustness when processing single-image inputs.

To validate our method, we collected RealCAD, a real-world dataset pairing CAD commands with multi-view images, captured freely on CAD models fabricated using 3D printing technology with various materials and textures.

Our contributions are: 
(a) We introduce CADCrafter, a latent diffusion-based framework to generate parametric CAD models from unconstrained images, leveraging geometric features to mitigate the domain gap between synthetic training data and in-the-wild testing data.
(b) We introduce an automatic code checker to learn CAD geometry validity by fine-tuning our diffusion model with direct preference optimization (DPO) thereby improving accuracy and reducing invalid outputs. %
(c) Our proposed CADCrafter framework accommodates both single-view and multi-view inputs. In addition, we introduce a dataset of unconstrained 3D printed CAD images paired with CAD commands, demonstrating the robustness and generalizability of the model.

\section{Related work}

\textbf{Generative models for CAD.} Most existing CAD generation research focuses on unconditional generation~\cite{DBLP:journals/tog/XuLJWWF24}  or conditional generation based on complete 3D information, such as point clouds~\cite{wu2021deepcad, ma2024draw}, sketches~\cite{li2020sketch2cad, wang2024vq}, B-reps~\cite{willis2021engineering, xu2021inferring} and voxel grids~\cite{li2023secad, li2024sfmcad}. For instance, DeepCAD~
\cite{wu2021deepcad} utilizes an autoencoder to encode CAD models and employs GANs for unconditional generation. SkexGen~\cite{xu2022skexgen} introduces an autoregressive generative model that encodes CAD construction sequences into disentangled codebooks. HNC-CAD~\cite{xu2023hierarchical} represents CAD models as a hierarchical tree of three levels of neural codes. Draw Step by Step~\cite{ma2024draw} incorporates a tokenizer to compress CAD point clouds and trains a multi-modal diffusion model for point cloud-conditioned generation. CAD-SIGNet~\cite{khan2024cad} proposes a layer-wise cross-attention mechanism between point clouds and CAD sequence embeddings. MultiCAD~\cite{ma2023multicad} develops a multimodal contrastive learning strategy to align CAD sequences with point clouds. More recently, Text2CAD~\cite{khan2024text2cad} has been introduced to generate parametric CAD models from text instructions.  Img2CAD~\cite{chen2024img2cad} is able to generate CAD command sequences with image inputs, however, it adopts a discriminative framework which results in limited performance \cite{chen2022utc}, especially on real-world objects.

Current CAD construction sequence data sets like DeepCAD~\cite{wu2021deepcad} and Fusion360~\cite{willis2021fusion} are standard synthetic data sets. These models are typically trained and tested on noise-free synthetic data. In this paper, we present a dataset that pairs multi-view images with CAD sequences and explores a more user-friendly approach by training only on synthetic data and evaluating our model on both synthetic and real-world captured data.

\noindent\textbf{3D Generative Models.} With the rise of large-scale model training, recent advances in 3D generative models have been significant. Most existing methods generate 3D shapes in discrete forms, such as implicit neural fields~\cite{poole2022dreamfusion, hong2023lrm, chen2024sculpt3d, chou2023diffusion}, point clouds~\cite{yang2019pointflow}, and meshes~\cite{wei2023taps3d, liu2024meshformer, siddiqui2024meshgpt, shen2024hmr}. However, these generated shapes often suffer from a lack of sharp geometric features and are not directly editable by users.

In contrast to these works, our model directly outputs sequences of CAD operations that can be readily imported into any CAD tool~\cite{AutoCAD, Fusion360, Onshape} for user editing.

\section{Approach} 
In this chapter, we present CADCrafter, a latent diffusion-based transformer tailored for generating CAD command sequences from images. Trained on a synthetic dataset and evaluated on both synthetic and real-world data, the model incorporates a geometry conditioning encoder to enhance geometric understanding and generalization. Additionally, we have developed a multi-view to single-view distillation technique to improve robustness for single-view inputs and introduced an annotation-free direct preference optimization method to improve accuracy in CAD representations.

\subsection{CAD Command Sequence Encoding}

The comprehensive CAD toolkit features an extensive array of commands, yet only a limited subset is frequently utilized in practice. Drawing on previous research \cite{willis2021fusion, wu2021deepcad}, we focus on two commonly used categories: \emph{sketch} and \emph{extrusion}, which offer ample expressive capabilities. We present an example of a simple CAD command sequence in Figure \ref{fig:teaser}. For simplicity, in \emph{sketch}, we adopt commands$\{\cmdSOL, \cmdLine, \cmdArc, \cmdCirc\}$, namely $\mathtt{start}$, $\mathtt{line}$, $\mathtt{arc}$, and $\mathtt{circle}$, to draw curves forming enclosed 2D regions named \emph{profile}. Then, each 2D \emph{profile} can be lifted to a 3D body using the \emph{extrusion} command $\cmdExt$. Each of these discrete commands is defined by its unique continuous parameters, which determine the size, location, scale, and type. 

Following the approach in DeepCAD \cite{wu2021deepcad}, we normalize all CAD models and quantize the continuous parameters into 256 levels represented as 8-bit integers to process the discrete and continuous command sequences. The $i$-th line of command $C_i$ is represented by the one-hot encoding of command types $s_i$ and a stacking of all parameters for all command types into a vector $p_i$, setting the unused parameters to $-1$. We then pad the sequence to a fixed length, $\cmdNum$, using the empty command $\cmdEOS$. To process CAD commands in a manner similar to natural language processing \cite{vaswani2017attention}, we tokenize the commands by mapping them to embedding spaces, the resulting embedding $e(C_i) = \embCmd+\embPm+\embPos\in\mathbb{R}^{\embDim}$, where $\embPos$ is a readable positional embedding and $\embDim=256$ is the embedding dimension. More details on the CAD commands and the tokenization process are provided in the supplementary. %

To facilitate the generation of sequential CAD data conditioned on images, as depicted in Figure \ref{fig:pipeline}, we initially train a transformer-based autoencoder to encode CAD sequences into latent vectors \(\z\) and subsequently reconstruct them. Then we adopt the latent diffusion framework \cite{rombach2022high}, which allows efficient learning and sampling within the latent CAD space with image conditioning. The design and training details of the autoencoder are similar to DeepCAD \cite{wu2021deepcad} and are included in the supplementary materials. %

\begin{figure}[t]
  \centering
   \vspace{5pt}
   \includegraphics[width=\linewidth]{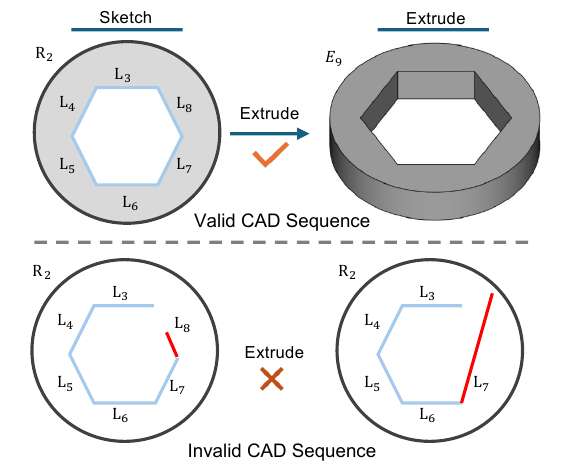}
 \vspace{-5pt}
   \caption{The code checker checks if the generated CAD command sequence is compilable. The first row illustrates cases that can be successfully compiled while the second row shows invalid cases where no 2D \emph{profile} is enclosed by the curves. The compiler inherently performs as an automatic checker to help our DPO fine-tuning process.}
   \label{fig:code_checker}
    \vspace{-20pt}
\end{figure}

\subsection{Geometry Conditioning Encoder}

CAD commands are precise operations based on geometry structures; thus, it is important to explore more geometric information from the input image. Therefore, we extract depth and surface normal maps with geometry estimation models \cite{yin2023metric3d}. Additionally, depth and normal are invariant to textures, which significantly reduces the domain gap between our texture-less synthetic renderings used for training and the unconstrained images used for testing.

Specifically, we feed the extracted depth and normal maps to the pre-trained DINO-V2\cite{oquab2023dinov2} encoder to get DINO features $h_i^{\text{depth}}$, $h_i^{\text{normal}} \in\mathbb{R}^{d_\text{dino}}$ where $i \in \{0,1,2,3\}$ indicates different views and $d_\text{dino} = 1536$. 

We designed a transformer-based geometry encoder that adaptively consolidates geometric cues from each view and modality, as each provides unique geometric information about the object. To help the model more effectively learn to integrate information from different modalities, the DINO features are stacked as patches and add a learnable modality embedding $e$ to get:
\begin{equation}
    \mathbf{H}=\mathit{cat}_{i=0}^3 (h_i^{\text{depth}}+e^{\text{depth}}, h_i^{\text{normal}}+e^{\text{normal}}),
\end{equation}
where $\mathbf{H} \in \mathbb{R}^{8 \times d_{\text{dino}}}$ and $\mathit{cat}$ denotes the concatenation operation across all views and modalities. We further apply a rotary positional embedding \cite{su2024roformer} to each token to help the multi-view geometry encoder effectively combine information. The averaged output feature $f_m$ of the geometry encoder is used as the conditioning vector.

\subsection{Denoising CAD Latent Vectors}
Since our sample space is the latent vectors \( z \) generated by the transformer-based autoencoder,  unlike conventional DDPM architectures that employ UNet structures tailored for image processing, we implement a diffusion transformer architecture \( \Omega \) to denoise the latent vector. The diffusion transformer architecture is similar to that of DALL-E 2 \cite{ramesh2022hierarchical}, which consists of layers with attention mechanisms \cite{vaswani2017attention}, fully connected layers, and layer normalization. 

Given the sampled CAD latent \( \z \), at each iteration, we add noise corresponding to a random timestep \( t \) to it to obtain \( \z_t \). The model then learns to restore the original latent \( \z_0 \). The diffusion model takes \( \z_t \),  \( f_m \) and \( \gamma(t) \) as inputs, where \( \gamma(t) \) is a positional embedding of timestep \( t \).

In our experiments, we found that directly predicting the original \( \z_0 \) yields better performance than predicting the added noise. Therefore, the loss function is defined as:
\begin{equation}
    \mathcal{L}_{\text{diff}} = \| \Omega(\z_t, \gamma(t)|f_m) - \z_0 \|^2.
\end{equation}
 During testing, we begin with a randomly sampled noise vector \( \z_T \sim \mathcal{N}(0, I) \) and iteratively apply our diffusion model to denoise it, ultimately producing the final output \( \z_0 \). The generated latents \( \z_0 \) are passed to the previously trained decoder to obtain the reconstructed CAD sequence.

\subsection{Multi-View to Single-View Distillation}

Creating a 3D object from a single-view input introduces inherent ambiguities, as the model must infer details from unseen areas. Instead of training a separate model for the single-view setting, we adopt the model trained from multi-view and distill the knowledge from the multi-view geometry encoder to the single-view geometry encoder by implicitly reducing the distance between the condition features.

Specifically, in single-view training, we freeze the weights of the trained multi-view geometry encoder as the reference model. We continue to feed multi-view inputs and employ a distillation loss to distill knowledge from it:
\vspace{-5pt}
\begin{equation}
    \mathcal{L}_{\text{distill}} = 1 - \frac{f_s \cdot f_m}{\|f_s\| \|f_m\|},
\end{equation}
where $f_m$ is the averaged output feature of the multi-view geometry encoder and $f_s$ is the averaged output feature of the single-view geometry encoder. The single-view geometry encoder is updated with $\mathcal{L}_{\text{distill}}$ and $ \mathcal{L}_{\text{diff}}$.

\subsection{Direct Preference Optimization based CAD Code Checker}

During training, diffusion loss concentrates on aligning distributions without explicit geometric supervision. However, CAD compilers enforce strict command rules. As shown in Figure~\ref{fig:code_checker}, there are instances where curves do not form closed surfaces, causing the generated CAD code to fail compiler checks.

To improve geometric precision, we take inspiration from reinforcement learning from human feedback (RLHF) \cite{bai2022training} in large language models (LLMs), where a reward function is trained from comparison data on model output to represent human preferences, and reinforcement learning is used to align the policy model.  Our key idea is to introduce a code checker to serve as an implicit reward model for the denoised latent code, utilizing direct preference optimization (DPO) \cite{rafailov2024direct, wallace2024diffusion} to fine-tune our approach.

Specifically, we generate multiple latent \( z \) vectors and use a CAD compiler to verify the compilability of the decoded commands. This automatic process enables us to pick a set of valid latent vectors (positive set) and a set of invalid latent vectors (negative set). We then fine-tune the diffusion model using the DPO loss, which is defined as follows:

\vspace{-10pt}
\begin{multline}
    L(\theta)
    = - \mathbb{E}_{
    (\zw_0, \zl_0) \sim \calD, t\sim \mathcal{U}(0,T), 
    \zw_{t}\sim q(\zw_{t}|\zw_0),\zl_{t} \sim q(\zl_{t}|\zl_0)
    } \\
    \log\sigma \left(-\frac{\beta}{2}  \left( \right. \right.
    \| \noise^w -\noise_\theta(\z_{t}^w,t)\|^2_2 - \|\noise^w - \noise_\text{ref}(\z_{t}^w,t)\|^2_2 \\
    \left. \left.  - \left( \| \noise^l -\noise_\theta(\z_{t}^l,t)\|^2_2 - \|\noise^l - \noise_\text{ref}(\z_{t}^l,t)\|^2_2\right)
    \right)\right)\label{eq:loss-dpo-1}.
\end{multline}

where $\epsilon$ is the noise in diffusion process, the loss term $l_w = \| \epsilon_w - \epsilon_\theta(\z_t^w, t) \|$
represents the model preference toward the positive sample \( \z_w \), while
$ l_l = \| \epsilon_l - \epsilon_{\text{ref}}(\z_t^l, t) \|$
represents the model's preference toward the negative sample \( \z_l \). It can observed that when optimizing the DPO loss, \( l_w \) decreases while \( l_l \) increases.  This adjustment increases the probability of generating positive samples and decreases the probability of producing code that fails the checks.

Furthermore, \( \epsilon_{\text{ref}} () \) denotes the frozen pretrained diffusion model trained in the second phase, and \( \epsilon_{\theta} () \) is the updating network. By limiting the difference between the finetuned model's output and the pre-trained model,  the effective knowledge acquired during pretraining can be preserved. The parameter \( \beta \) controls the regularization of the fine-tuned model's distance from the original model. A larger \( \beta \) imposes more constraints when the model deviates from the pre-trained model. In our experiments, we set \( \beta = 20 \).

\section{Experiments} 

\begin{figure}[b]
\vspace{-10pt}
  \centering
   \includegraphics[width=\linewidth]{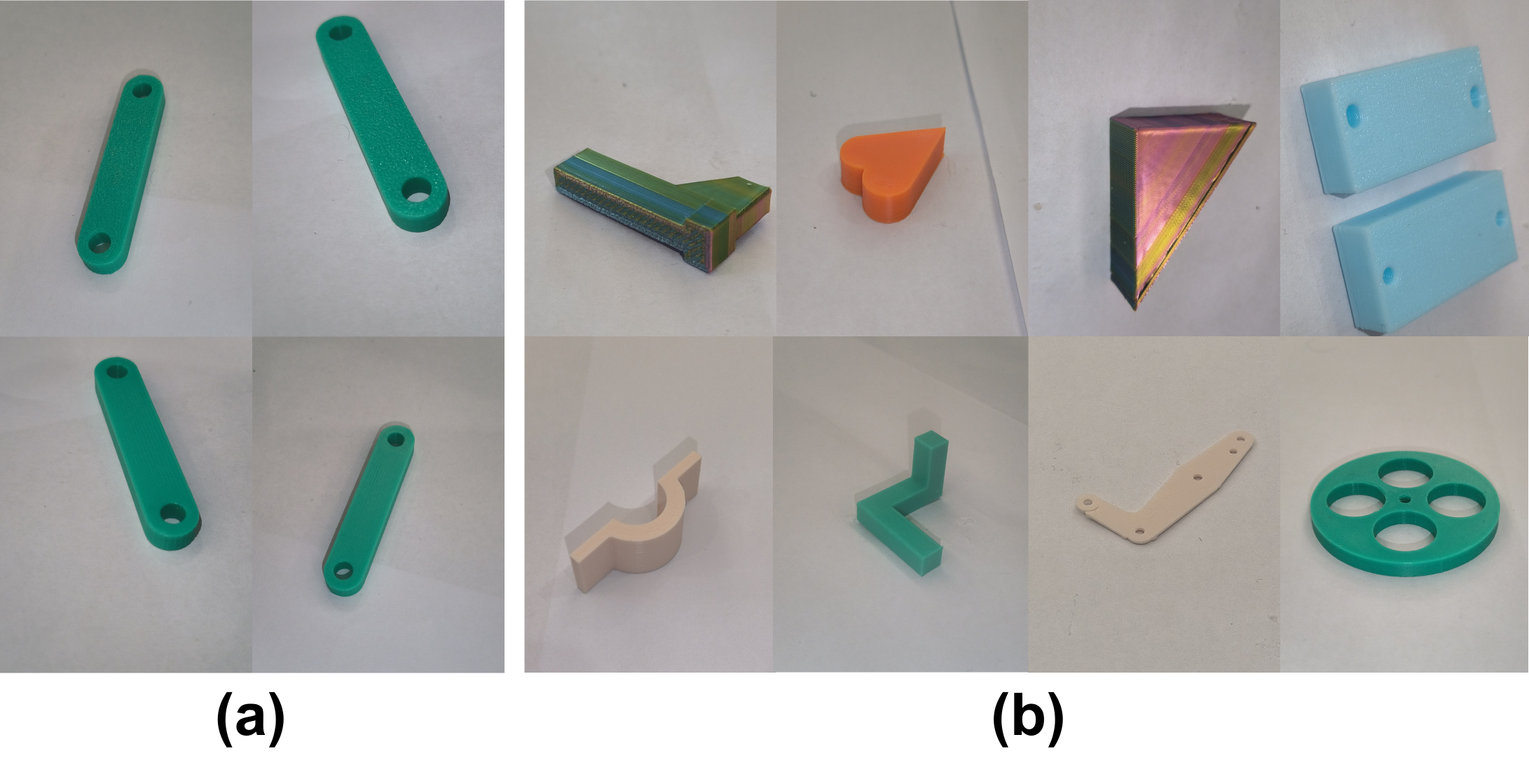}
\vspace{-20pt}
   \caption{We showcase our RealCAD dataset: (a) casually captured multi-view images of a 3D printed CAD model, (b) more examples of 3D printed CAD models freely captured with iPhones.}
   \label{fig:dataset}
\end{figure}

\subsection{Experimental Setups}
\textbf{Datasets}. We train our method solely on \textbf{DeepCAD} \cite{wu2021deepcad} training set, a dataset composed mainly of CAD mechanical parts. We render 8 sets of 4-view images around the CAD model with a random elevation and perturbations on azimuth. For the single-view setting, we randomly pick one image for each training step. 

 To better assess the generalizability of the model, we collect our own real-world testing dataset \textbf{RealCAD}. We randomly select 150 CAD models from the DeepCAD test set and fabricate them with 3D printing using various textures and materials. As shown in Figure \ref{fig:dataset}, we casually take 4-view images around the printed objects without specific requirements.

\textbf{Implementation Details.} All experiments are performed on a single RTX6000 Ada GPU. In the first stage, we train the autoencoder for 1,000 epochs using a learning rate of $2 \times 10^{-4}$. In the second stage, we train the image-conditioned diffusion model with a batch size of 2,048 for 3,000 epochs using a learning rate of $5 \times 10^{-5}$. In the DPO finetuning stage, we collect 10000 pairs of positive and negative pairs and use them to further train the diffusion model for 500 epochs. We uniformly use the pre-trained Metric3D~\cite{yin2023metric3d} as depth and normal extractors.

\textbf{Evaluation Metrics.} Following previous work \cite{wu2021deepcad, ma2024draw}, we adopt Command Accuracy ($Acc_{cmd}$ in $\%$), which measures the correctness of the predicted CAD command types, and Parameter Accuracy ($Acc_{para}$ in $\%$), which measures the correctness of the command parameters once the command type is correctly recovered. Both $Acc_{cmd}$ and $Acc_{para}$ assess how closely the reconstructed CAD sequences resemble the original human-designed sequences. The final CAD models are also quantitatively evaluated against the ground-truth CAD models using Median Chamfer Distance (Med CD), which measures the geometric similarity between the reconstructed and ground-truth models. Additionally, we use the Invalid Rate (IR) to evaluate the percentage of CAD sequences output that fail to compile.

\textbf{Baselines}. DeepCAD \cite{wu2021deepcad} provides a point cloud-conditioned generation scheme; therefore, we replace the point cloud encoder with a DINO-V2 image encoder and adaptive layers \cite{wu2021deepcad} to allow image-conditioned generation. HNC-CAD~\cite{xu2023hierarchical} supports the conditional generation of partial command CAD sequences, we retrain the networks by feeding the DINO image features to its original conditional encoder. However, since HNC-CAD directly generates explicit \emph{loop}, \emph{profile} and \emph{solids} instead of commands, we do not include comparisons for command accuracy ($Acc_{cmd}$) and parameter accuracy ($Acc_{para}$) with this model. Img2CAD \cite{chen2024img2cad} is capable of generating CAD commands from images. Since their code is not publicly available, we use the performance metrics reported directly in their paper.
Besides CAD generation methods, we also compare our approach with recent image-to-3D methods, such as One-2-3-45 \cite{liu2024one}, Wonder3D \cite{long2024wonder3d}, and TripoSR \cite{tochilkin2024triposr}. Wonder3D \cite{long2024wonder3d} also utilizes normal information for 3D reconstruction.  To ensure a fair comparison, we fine-tune One-2-3-45 \cite{liu2024one} and Wonder3D \cite{long2024wonder3d} using our rendered DeepCAD data, while TripoSR \cite{tochilkin2024triposr} is a commercial model without publicly accessible training codes. Since they directly generate 3D shapes, we only evaluate the Median Chamfer Distance (Med CD) against these methods. In subsequent sections, $^{\dagger}$ denotes models fine-tuned on our CAD data, while $^*$ indicates models for which we replaced the original condition encoder with an image encoder and then retrained.

\begin{figure*}
  \centering
   \includegraphics[width=\linewidth]{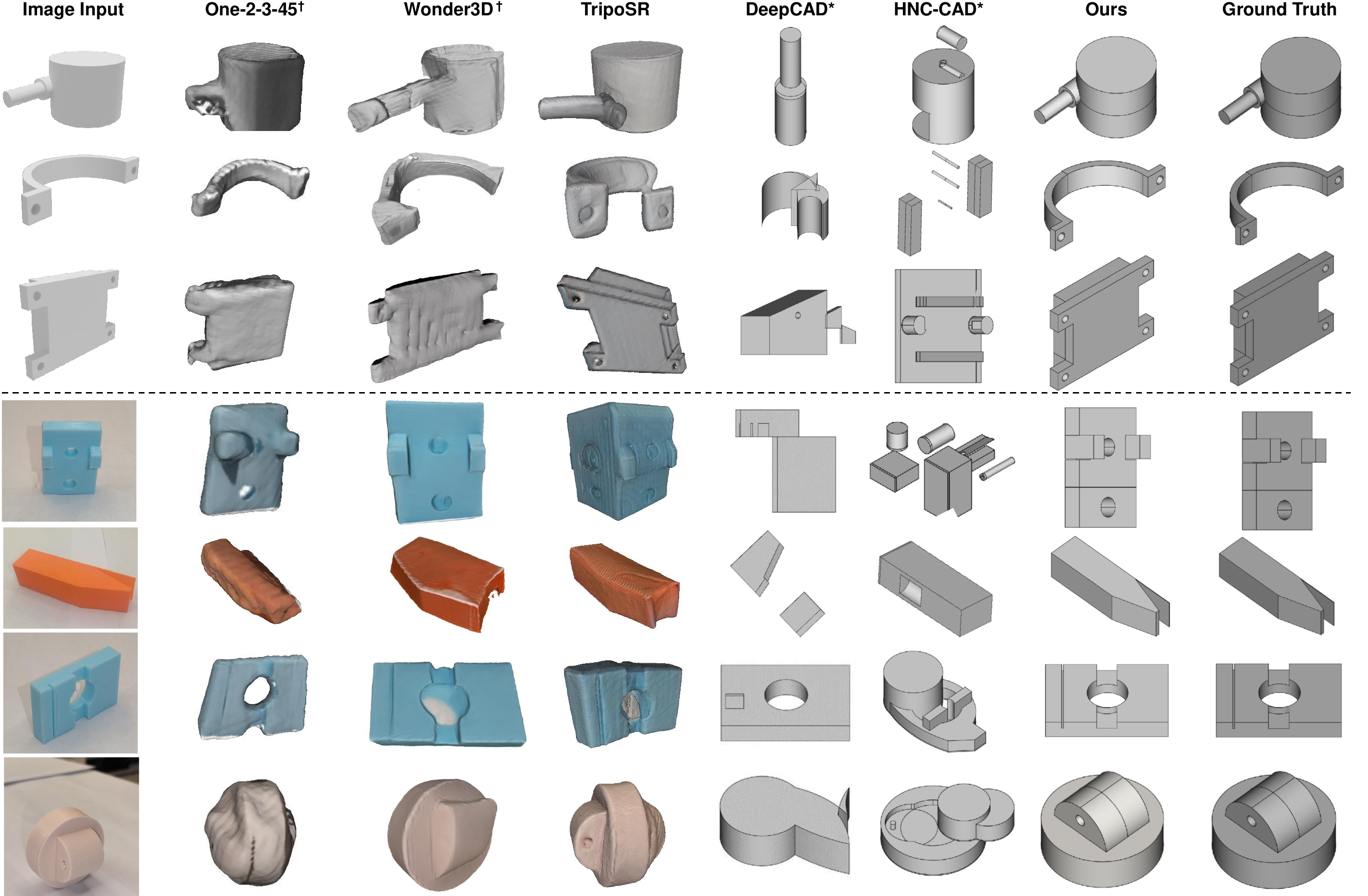}
\vspace{-15pt}
   \caption{We compare the generated CAD models from single-view images with existing methods on two datasets: the upper part shows results on the DeepCAD renderings, and the lower part shows results on the real-world RealCAD dataset. }
   \label{fig:rendered_results}
   \vspace{-15pt}
\end{figure*}

\begin{table}
\setlength{\tabcolsep}{1.0pt}
\fontsize{9}{10}\selectfont
\centering
\begin{tabular}{lcccc}
\toprule
\textbf{Methods} & $\text{ACC}_{\text{cmd}} \uparrow$ & $\text{ACC}_{\text{para}} \uparrow$ & \textbf{Med CD} $\downarrow$ & \textbf{IR} $\downarrow$ \\ 
\midrule
\multicolumn{5}{l}{\textbf{\textit{DeepCAD test set (synthetic)}}} \\ \midrule
DeepCAD$^*_{s}$ & 77.72 & 65.30 & 0.126 & 0.123 \\
DeepCAD$^*_{m}$ & 79.62 & 66.75 & 0.113 & 0.106 \\ 
HNC-CAD$^*_{s}$ & - & - & 0.214 & 0.114 \\
HNC-CAD$^*_{m}$ & - & - & 0.208 & 0.101 \\
Img2CAD  & 80.57 & 68.77 & 0.160 & 0.288  \\ 
\hdashline
TripoSR & - & - & 0.136 & - \\
One-2-3-45$^{\dagger}$ & - & - & 0.151 & -  \\
Wonder3D$^{\dagger}$ & - & - & 0.133 & -  \\
\midrule
$\mathbf{CADCrafter_{s}}$ & $\mathbf{83.23}$ & $\mathbf{71.82}$ & $\mathbf{0.049}$ & $\mathbf{0.042}$ \\
$\mathbf{CADCrafter_{m}}$ & $\mathbf{84.62}$ & $\mathbf{73.31}$ & $\mathbf{0.026}$ & $\mathbf{0.036}$ \\
\midrule
\multicolumn{5}{l}{\textbf{\textit{RealCAD Dataset (real-world)}}} \\ \midrule
DeepCAD$^*_{s}$ & 56.59 & 41.32 & 0.264 & 0.527 \\
DeepCAD$^*_{m}$ & 54.11 & 37.27 &  0.295 & 0.567 \\ 
HNC-CAD$^*_{s}$ & - & - & 0.276 & 0.147 \\
HNC-CAD$^*_{m}$ & - & - & 0.305 & 0.167 \\
\hdashline
TripoSR & - & - & 0.128 & - \\
One-2-3-45$^{\dagger}$ & - & - & 0.147 & -  \\
Wonder3D$^{\dagger}$ & - & - & 0.125 & -  \\
\midrule
$\mathbf{CADCrafter_{s}}$ & $\mathbf{81.23}$ & $\mathbf{64.16}$ & $\mathbf{0.082}$ & $\mathbf{0.087}$ \\
$\mathbf{CADCrafter_{m}}$ & $\mathbf{83.18}$ & $\mathbf{66.89}$ & $\mathbf{0.062}$ & $\mathbf{0.067}$ \\
\bottomrule
\end{tabular}
\vspace{-10pt}
\caption{Performance comparisons on the synthetic DeepCAD dataset and real-world RealCAD dataset where $s$ denotes single-view and $m$ denotes multi-view settings.} %
\label{tab-deepcad}
\vspace{-17pt}
\end{table}

\vspace{-3pt}
\subsection{Quantitative Results}
We compare our model with existing approaches in Table~\ref{tab-deepcad}. On DeepCAD \cite{wu2021deepcad} dataset, the results show that our CADCrafter achieves high CAD sequence accuracy and significantly reduces the failure rate in various scenarios on both multi-view and single-view tasks. Our method outperforms  DeepCAD \cite{wu2021deepcad}, HNC-CAD \cite{xu2023hierarchical} and Img2CAD \cite{chen2024img2cad} in each criterion, especially the invalid rate, demonstrating the robustness of our method. On RealCAD dataset, the performance of DeepCAD \cite{wu2021deepcad} and HNC-CAD \cite{xu2023hierarchical} significantly declined across all criteria, indicating their overfitting to synthetic data and inability to generalize to real-world scenarios. This underscores the significant domain gap between rendered synthetic data and real-world captured data. In contrast, our method, despite being trained solely on synthetic data, generalizes effectively to real-world data with only a slight performance drop, maintaining high accuracy and low invalid rates.

We also benchmarked our method against advanced large-scale image-to-3D generative models One-2-3-45 \cite{liu2024one}, Wonder3D \cite{long2024wonder3d}, and TripoSR \cite{tochilkin2024triposr}, which demonstrated consistent performance across both synthetic and real-world data due to their generalizability. However, our method consistently outperformed these models in terms of geometric accuracy in both scenarios.

\vspace{-3pt}
\subsection{Qualitative Results}
\vspace{-2pt}
We compare our method with existing methods qualitatively in Figure \ref{fig:rendered_results}. Our method successfully recovered the CAD command sequences from the single image input on both synthetic and real-world scenarios, while DeepCAD \cite{wu2021deepcad} and HNC-CAD \cite{xu2023hierarchical} failed to produce meaningful shapes in both cases. Wonder3D \cite{long2024wonder3d} and TripoSR \cite{tochilkin2024triposr} generate better results than One-2-3-45 \cite{liu2024one}. However, the generated shapes often exhibit unsmooth surfaces and lack precision. Additionally, they consistently fail to accurately reproduce standard geometric shapes such as rectangles and circles, making these shape approximations unsuitable for manufacturing applications.

\subsection{Ablation Studies of Different Modalities}

\begin{table}[b]
\setlength{\tabcolsep}{1.0pt}
\fontsize{9}{10}\selectfont
\centering
\vspace{-10pt}
\begin{tabular}{lccccccc}
\toprule
        \textbf{Inputs} & $\text{ACC}_{\text{cmd}} \uparrow$ & $\text{ACC}_{\text{para}} \uparrow$ & \textbf{Med CD} $\downarrow$ & \textbf{IR} $\downarrow$ \\ 
        \midrule
        \multicolumn{5}{l}{\textbf{\textit{DeepCAD test set (synthetic)}}} \\ \midrule
        RGB  & 83.26 & 72.09 & 0.029 & 0.037 \\
        Normal  & 83.14 & 72.63 & 0.032 & 0.039 \\
        Depth  & 83.06 & 72.28 & 0.036 & 0.041 \\
        RGB+Depth+Normal & \textbf{85.18} & \textbf{74.86} & \textbf{0.023} & \textbf{0.031} \\
        Depth+Normal (Ours) & 84.62 & 73.31 & 0.026 & 0.036 \\
        
        \midrule
        \multicolumn{5}{l}{\textbf{\textit{RealCAD Dataset (real-world)}}} \\ \midrule
        RGB & 77.92 & 49.73 & 0.219 & 0.26  \\ 
        Normal & 82.57 & 65.77 & 0.106 & 0.073  \\ 
        Depth & 82.17 & 64.78 & 0.102 & 0.087  \\ 
        RGB+Depth+Normal  & 78.67 & 52.78 & 0.192 & 0.227  \\
        Depth+Normal (Ours)  & \textbf{83.18} & \textbf{66.89} & \textbf{0.062} & \textbf{0.067}  \\ 
        
\midrule
\end{tabular}
\vspace{-10pt}
\caption{Ablation studies on various geometric modalities with multi-view inputs reveal that while RGB slightly enhances performance on synthetic data, it significantly reduces the model's generalizability.}
\label{Ablation_modality}
\vspace{-15pt}
\end{table}

To study the impact of different modalities, we sequentially train and test our model on different combinations of modalities or each modality alone. As shown in Table \ref{Ablation_modality}, the results show that since CAD models inherently lack textures, using solely rendered images, normals, or depth maps yields similar outcomes on the synthetic DeepCAD dataset. Each modality captures unique information: normals emphasize the relationships between surfaces, while depth maps focus on object scale. Thus, combining all three modalities leads to the best performance while tested on synthetic setting. 

However, testing the trained models on real-world images reveals that normals and depth maps, which focus solely on geometric characteristics, are not impacted by the domain gap introduced by the textures of the CAD model. While incorporating an RGB image as input can slightly improve performance on synthetic data, the significant difference between synthetic RGB data and real images markedly reduces the model's generalizability when trained with RGB inputs. Therefore, to enhance the model's generalizability, we exclude RGB images during both training and testing.

\subsection{Ablations of Different Components}
We perform ablation studies of different components in the single-view setting in Table \ref{components}.\\
\begin{table}\setlength{\tabcolsep}{1.0pt}
\fontsize{9}{10}\selectfont
\centering

\begin{tabular}{lccccccc}
\toprule
        \textbf{Methods} & $\text{ACC}_{\text{cmd}} \uparrow$ & $\text{ACC}_{\text{para}} \uparrow$ & \textbf{Med CD} $\downarrow$ & \textbf{IR} $\downarrow$ \\ 
        \midrule
        CADCrafter$_{w/o-L_{Geo}}$  & 81.89 & 69.98 & 0.056 & 0.059 \\
        CADCrafter$_{scratch}$  & 80.61 & 68.62 & 0.079 & 0.078 \\
        CADCrafter$_{w/o-L_{distill}}$  & 81.12 & 69.83 & 0.068 & 0.072 \\
        CADCrafter$_{w/o-L_{dpo}}$  & 81.64 & 69.32 & 0.072 & 0.081 \\
        CADCrafter  & \textbf{83.23} & \textbf{71.82} & \textbf{0.049} & \textbf{0.042} \\

\midrule
\end{tabular}
\vspace{-10pt}
\caption{Ablation studies of different components on the DeepCAD dataset with single-view inputs.}\label{table_Components}
\vspace{-0.2in}
\label{components}
\end{table}

\noindent\textbf{Geometric Encoder}. In CADCrafter$_{w/o-L_{Geo}}$, we replace our geometric encoder with concatenated DINO features, processed through a 3-layer MLP to produce the conditional embedding. This change led to a noticeable decrease in performance, demonstrating that our transformer-based geometric encoder effectively consolidates geometric information across different modalities.

\noindent\textbf{From Scratch}. In CADCrafter$_{scratch}$, we train the single-view geometry encoder and the diffusion model from scratch. The decline in performance indicates that our sequential training strategy, which leverages multi-view knowledge, benefits the single-view configuration.

\noindent\textbf{Multi-view distillation}. In CADCrafter$_{w/o-L_{distill}}$, we train the single-view geometry encoder alongside a diffusion model that was pre-trained in a multi-view setting but without employing our distillation loss. The performance decline confirms that the distillation loss effectively transfers comprehensive knowledge from the multi-view encoder to the single-view encoder.

\noindent\textbf{Direct Preference Optimization}. \\In CADCrafter$_{w/o-L_{dpo}}$, we remove the DPO fine-tuning alone. The results demonstrate that DPO effectively lowers the CAD code invalid rate (IR) and improves command accuracy. This underscores the effectiveness of our code checker in helping the model learn more accurate code patterns with geometric constraints.

 \begin{figure}[b]
  \centering
   \includegraphics[width=0.99\linewidth]{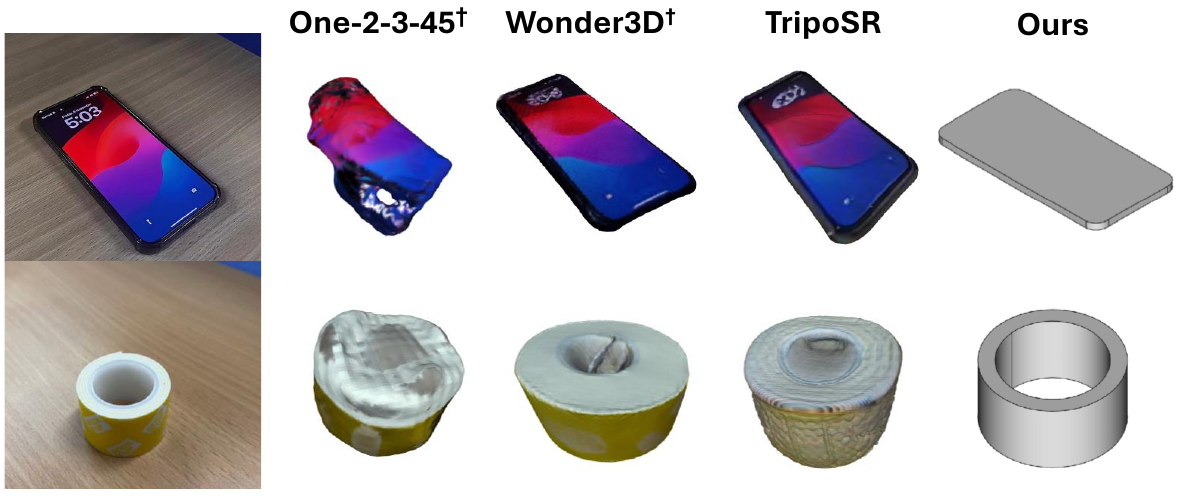}
\vspace{-10pt}
   \caption{We compare our image-to-CAD results on unseen general objects with image-to-3d methods.}
   \label{fig:general_results}
\end{figure}

\subsection{Applications of CADCrafter}
Creating digital twins with CAD models largely benefits the manufacturing industry and Embodied AI simulation. Currently, we train exclusively on synthetic datasets of mechanical parts but have successfully demonstrated the conversion of casually captured real-world objects into editable CAD models, as illustrated in Figure \ref{fig:general_results}. To our knowledge, we are the first to showcase this capability. While the current generation complexity is limited by existing datasets, advancements in the field should enable the conversion of more complex objects into precise CAD models.

\begin{figure}[t]
  \centering
   \includegraphics[width=\linewidth]{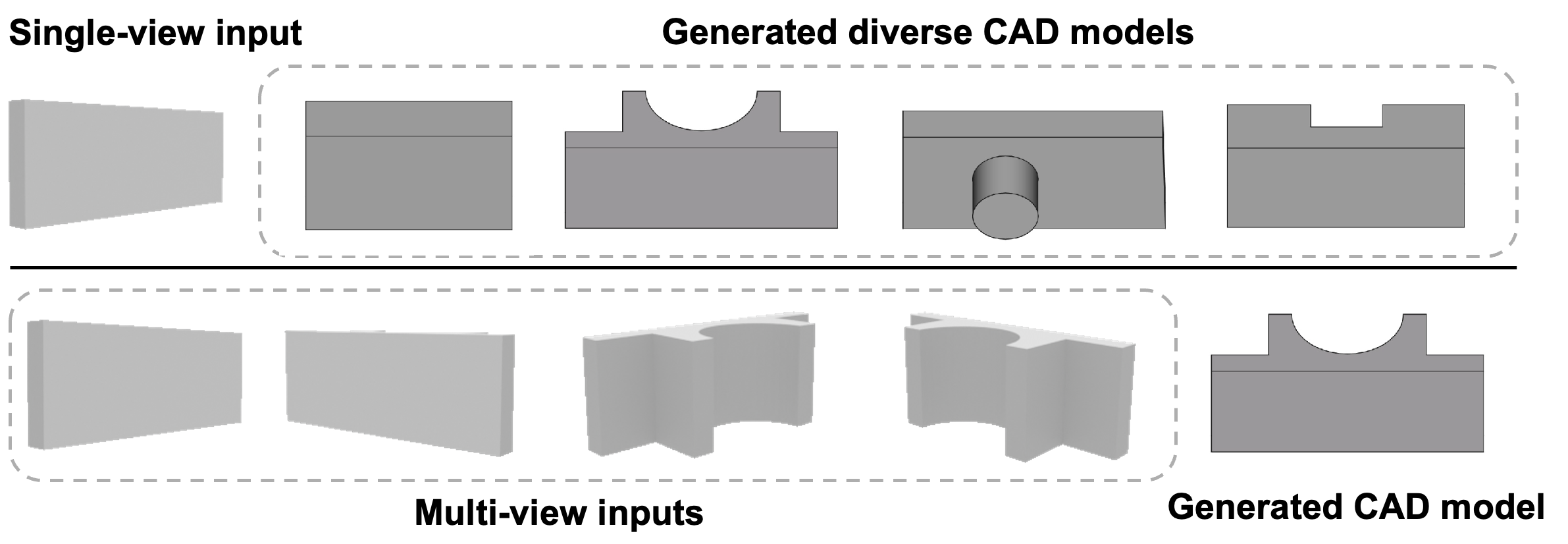}
\vspace{-10pt}
   \caption{In the single-view setting, CADCrafter can generate diverse shapes for the unseen parts. Given the multi-view input, CADCrafter is able to reconstruct more accurate shapes.}
   \label{fig:single_multi_view}
   \vspace{-10pt}
\end{figure}

Since the single-view images cannot capture the complete information of the objects, it is desirable to provide various choices given the partial observations. In the first row of Figure \ref{fig:single_multi_view}, we demonstrate that our model can offer various CAD models with different unseen parts in the single-view image. Users can further choose and edit these generated results to suit their specific needs. Additionally, as illustrated in the second row of Figure \ref{fig:single_multi_view}, multi-view images provide more sufficient information about the shape geometry, and our model can generate more specific models when precise results are needed.

\section{Conclusions and Future Work}
We introduce CADCrafter, a latent diffusion model that converts images into CAD command sequences using geometric information. Trained solely on synthetic data, CADCrafter generalizes effectively to real-world images and unseen object types. Our geometric encoder bridges synthetic-real domain gaps by capturing diverse shape information and distilling multi-view knowledge into a single-view encoder, enhancing single-view performance. Additionally, we propose an automated code checker using direct preference optimization to incorporate CAD compiler feedback, improving geometric accuracy. We also contribute a new dataset of unconstrained images of 3D-printed CAD models with corresponding commands for validation. Future work includes incorporating physical properties essential for manufacturing and extending the model with text-based editing capabilities.

\section*{Acknowledgements} 
\vspace{-5pt} 
This research work is supported by the Agency for Science, Technology and Research (A*STAR) under its MTC Programmatic Funds (Grant No. M23L7b0021). %

{
    \small
    \bibliographystyle{ieeenat_fullname}
    \bibliography{main}

\begin{thebibliography}{45}
\providecommand{\natexlab}[1]{#1}
\providecommand{\url}[1]{\texttt{#1}}
\expandafter\ifx\csname urlstyle\endcsname\relax
  \providecommand{\doi}[1]{doi: #1}\else
  \providecommand{\doi}{doi: \begingroup \urlstyle{rm}\Url}\fi

\bibitem[Aut()]{AutoCAD}
Autocad.
\newblock \url{https://www.autodesk.com/products/autocad}.

\bibitem[Fus()]{Fusion360}
Fusion 360.
\newblock \url{https://www.autodesk.com/products/fusion-360}.

\bibitem[Ons()]{Onshape}
Onshape.
\newblock \url{http://onshape.com}.

\bibitem[Bai et~al.(2022)Bai, Jones, Ndousse, Askell, Chen, DasSarma, Drain, Fort, Ganguli, Henighan, et~al.]{bai2022training}
Yuntao Bai, Andy Jones, Kamal Ndousse, Amanda Askell, Anna Chen, Nova DasSarma, Dawn Drain, Stanislav Fort, Deep Ganguli, Tom Henighan, et~al.
\newblock Training a helpful and harmless assistant with reinforcement learning from human feedback.
\newblock \emph{arXiv preprint arXiv:2204.05862}, 2022.

\bibitem[Chen et~al.(2022)Chen, Tan, Cheng, Jiang, Liu, Zhu, and Gu]{chen2022utc}
Cheng Chen, Zhenshan Tan, Qingrong Cheng, Xin Jiang, Qun Liu, Yudong Zhu, and Xiaodong Gu.
\newblock Utc: A unified transformer with inter-task contrastive learning for visual dialog.
\newblock In \emph{Proceedings of the IEEE/CVF Conference on computer vision and pattern recognition}, pages 18103--18112, 2022.

\bibitem[Chen et~al.(2024{\natexlab{a}})Chen, Yang, Yang, Feng, Fu, Foo, Lin, and Liu]{chen2024sculpt3d}
Cheng Chen, Xiaofeng Yang, Fan Yang, Chengzeng Feng, Zhoujie Fu, Chuan-Sheng Foo, Guosheng Lin, and Fayao Liu.
\newblock Sculpt3d: Multi-view consistent text-to-3d generation with sparse 3d prior.
\newblock In \emph{Proceedings of the IEEE/CVF Conference on Computer Vision and Pattern Recognition}, pages 10228--10237, 2024{\natexlab{a}}.

\bibitem[Chen et~al.(2024{\natexlab{b}})Chen, Yu, Hu, Li, Xu, Cao, Zhu, Zang, Zhang, Li, et~al.]{chen2024img2cad}
Tianrun Chen, Chunan Yu, Yuanqi Hu, Jing Li, Tao Xu, Runlong Cao, Lanyun Zhu, Ying Zang, Yong Zhang, Zejian Li, et~al.
\newblock Img2cad: Conditioned 3d cad model generation from single image with structured visual geometry.
\newblock \emph{arXiv preprint arXiv:2410.03417}, 2024{\natexlab{b}}.

\bibitem[Chou et~al.(2023)Chou, Bahat, and Heide]{chou2023diffusion}
Gene Chou, Yuval Bahat, and Felix Heide.
\newblock Diffusion-sdf: Conditional generative modeling of signed distance functions.
\newblock In \emph{Proceedings of the IEEE/CVF international conference on computer vision}, pages 2262--2272, 2023.

\bibitem[Dong et~al.(2024)Dong, Zuo, Gu, Yuan, Zhao, Dong, Bo, and Huang]{DBLP:conf/cvpr/DongZ0YZDBH24}
Yuan Dong, Qi Zuo, Xiaodong Gu, Weihao Yuan, Zhengyi Zhao, Zilong Dong, Liefeng Bo, and Qixing Huang.
\newblock {GPLD3D:} latent diffusion of 3d shape generative models by enforcing geometric and physical priors.
\newblock In \emph{{IEEE/CVF} Conference on Computer Vision and Pattern Recognition, {CVPR} 2024, Seattle, WA, USA, June 16-22, 2024}, pages 56--66. {IEEE}, 2024.

\bibitem[Ho et~al.(2020)Ho, Jain, and Abbeel]{ho2020denoising}
Jonathan Ho, Ajay Jain, and Pieter Abbeel.
\newblock Denoising diffusion probabilistic models.
\newblock \emph{Advances in neural information processing systems}, 33:\penalty0 6840--6851, 2020.

\bibitem[Hong et~al.(2023)Hong, Zhang, Gu, Bi, Zhou, Liu, Liu, Sunkavalli, Bui, and Tan]{hong2023lrm}
Yicong Hong, Kai Zhang, Jiuxiang Gu, Sai Bi, Yang Zhou, Difan Liu, Feng Liu, Kalyan Sunkavalli, Trung Bui, and Hao Tan.
\newblock Lrm: Large reconstruction model for single image to 3d.
\newblock \emph{arXiv preprint arXiv:2311.04400}, 2023.

\bibitem[Khan et~al.(2024{\natexlab{a}})Khan, Dupont, Ali, Cherenkova, Kacem, and Aouada]{khan2024cad}
Mohammad~Sadil Khan, Elona Dupont, Sk~Aziz Ali, Kseniya Cherenkova, Anis Kacem, and Djamila Aouada.
\newblock Cad-signet: Cad language inference from point clouds using layer-wise sketch instance guided attention.
\newblock In \emph{Proceedings of the IEEE/CVF Conference on Computer Vision and Pattern Recognition}, pages 4713--4722, 2024{\natexlab{a}}.

\bibitem[Khan et~al.(2024{\natexlab{b}})Khan, Sinha, Sheikh, Stricker, Ali, and Afzal]{khan2024text2cad}
Mohammad~Sadil Khan, Sankalp Sinha, Talha~Uddin Sheikh, Didier Stricker, Sk~Aziz Ali, and Muhammad~Zeshan Afzal.
\newblock Text2cad: Generating sequential cad models from beginner-to-expert level text prompts.
\newblock \emph{arXiv preprint arXiv:2409.17106}, 2024{\natexlab{b}}.

\bibitem[Li et~al.(2020)Li, Pan, Bousseau, and Mitra]{li2020sketch2cad}
Changjian Li, Hao Pan, Adrien Bousseau, and Niloy~J Mitra.
\newblock Sketch2cad: Sequential cad modeling by sketching in context.
\newblock \emph{ACM Transactions on Graphics (TOG)}, 39\penalty0 (6):\penalty0 1--14, 2020.

\bibitem[Li et~al.(2023)Li, Guo, Zhang, and Yan]{li2023secad}
Pu Li, Jianwei Guo, Xiaopeng Zhang, and Dong-Ming Yan.
\newblock Secad-net: Self-supervised cad reconstruction by learning sketch-extrude operations.
\newblock In \emph{Proceedings of the IEEE/CVF Conference on Computer Vision and Pattern Recognition}, pages 16816--16826, 2023.

\bibitem[Li et~al.(2024)Li, Guo, Li, Benes, and Yan]{li2024sfmcad}
Pu Li, Jianwei Guo, Huibin Li, Bedrich Benes, and Dong-Ming Yan.
\newblock Sfmcad: Unsupervised cad reconstruction by learning sketch-based feature modeling operations.
\newblock In \emph{Proceedings of the IEEE/CVF Conference on Computer Vision and Pattern Recognition}, pages 4671--4680, 2024.

\bibitem[Liu et~al.(2024{\natexlab{a}})Liu, Xu, Jin, Chen, Varma~T, Xu, and Su]{liu2024one}
Minghua Liu, Chao Xu, Haian Jin, Linghao Chen, Mukund Varma~T, Zexiang Xu, and Hao Su.
\newblock One-2-3-45: Any single image to 3d mesh in 45 seconds without per-shape optimization.
\newblock \emph{Advances in Neural Information Processing Systems}, 36, 2024{\natexlab{a}}.

\bibitem[Liu et~al.(2024{\natexlab{b}})Liu, Zeng, Wei, Shi, Chen, Xu, Zhang, Wang, Zhang, Liu, et~al.]{liu2024meshformer}
Minghua Liu, Chong Zeng, Xinyue Wei, Ruoxi Shi, Linghao Chen, Chao Xu, Mengqi Zhang, Zhaoning Wang, Xiaoshuai Zhang, Isabella Liu, et~al.
\newblock Meshformer: High-quality mesh generation with 3d-guided reconstruction model.
\newblock \emph{arXiv preprint arXiv:2408.10198}, 2024{\natexlab{b}}.

\bibitem[Liu et~al.(2023)Liu, Wu, Van~Hoorick, Tokmakov, Zakharov, and Vondrick]{liu2023zero}
Ruoshi Liu, Rundi Wu, Basile Van~Hoorick, Pavel Tokmakov, Sergey Zakharov, and Carl Vondrick.
\newblock Zero-1-to-3: Zero-shot one image to 3d object.
\newblock In \emph{Proceedings of the IEEE/CVF international conference on computer vision}, pages 9298--9309, 2023.

\bibitem[Long et~al.(2024)Long, Guo, Lin, Liu, Dou, Liu, Ma, Zhang, Habermann, Theobalt, et~al.]{long2024wonder3d}
Xiaoxiao Long, Yuan-Chen Guo, Cheng Lin, Yuan Liu, Zhiyang Dou, Lingjie Liu, Yuexin Ma, Song-Hai Zhang, Marc Habermann, Christian Theobalt, et~al.
\newblock Wonder3d: Single image to 3d using cross-domain diffusion.
\newblock In \emph{Proceedings of the IEEE/CVF Conference on Computer Vision and Pattern Recognition}, pages 9970--9980, 2024.

\bibitem[Ma et~al.(2023)Ma, Xu, Li, and Zhou]{ma2023multicad}
Weijian Ma, Minyang Xu, Xueyang Li, and Xiangdong Zhou.
\newblock Multicad: Contrastive representation learning for multi-modal 3d computer-aided design models.
\newblock In \emph{Proceedings of the 32nd ACM International Conference on Information and Knowledge Management}, pages 1766--1776, 2023.

\bibitem[Ma et~al.(2024)Ma, Chen, Lou, Li, and Zhou]{ma2024draw}
Weijian Ma, Shuaiqi Chen, Yunzhong Lou, Xueyang Li, and Xiangdong Zhou.
\newblock Draw step by step: Reconstructing cad construction sequences from point clouds via multimodal diffusion.
\newblock In \emph{Proceedings of the IEEE/CVF Conference on Computer Vision and Pattern Recognition}, pages 27154--27163, 2024.

\bibitem[Oquab et~al.(2023)Oquab, Darcet, Moutakanni, Vo, Szafraniec, Khalidov, Fernandez, Haziza, Massa, El-Nouby, Howes, Huang, Xu, Sharma, Li, Galuba, Rabbat, Assran, Ballas, Synnaeve, Misra, Jegou, Mairal, Labatut, Joulin, and Bojanowski]{oquab2023dinov2}
Maxime Oquab, Timothée Darcet, Theo Moutakanni, Huy~V. Vo, Marc Szafraniec, Vasil Khalidov, Pierre Fernandez, Daniel Haziza, Francisco Massa, Alaaeldin El-Nouby, Russell Howes, Po-Yao Huang, Hu Xu, Vasu Sharma, Shang-Wen Li, Wojciech Galuba, Mike Rabbat, Mido Assran, Nicolas Ballas, Gabriel Synnaeve, Ishan Misra, Herve Jegou, Julien Mairal, Patrick Labatut, Armand Joulin, and Piotr Bojanowski.
\newblock Dinov2: Learning robust visual features without supervision, 2023.

\bibitem[Poole et~al.(2022)Poole, Jain, Barron, and Mildenhall]{poole2022dreamfusion}
Ben Poole, Ajay Jain, Jonathan~T Barron, and Ben Mildenhall.
\newblock Dreamfusion: Text-to-3d using 2d diffusion.
\newblock \emph{arXiv preprint arXiv:2209.14988}, 2022.

\bibitem[Rafailov et~al.(2024)Rafailov, Sharma, Mitchell, Manning, Ermon, and Finn]{rafailov2024direct}
Rafael Rafailov, Archit Sharma, Eric Mitchell, Christopher~D Manning, Stefano Ermon, and Chelsea Finn.
\newblock Direct preference optimization: Your language model is secretly a reward model.
\newblock \emph{Advances in Neural Information Processing Systems}, 36, 2024.

\bibitem[Ramesh et~al.(2022)Ramesh, Dhariwal, Nichol, Chu, and Chen]{ramesh2022hierarchical}
Aditya Ramesh, Prafulla Dhariwal, Alex Nichol, Casey Chu, and Mark Chen.
\newblock Hierarchical text-conditional image generation with clip latents.
\newblock \emph{arXiv preprint arXiv:2204.06125}, 1\penalty0 (2):\penalty0 3, 2022.

\bibitem[Rombach et~al.(2022)Rombach, Blattmann, Lorenz, Esser, and Ommer]{rombach2022high}
Robin Rombach, Andreas Blattmann, Dominik Lorenz, Patrick Esser, and Bj{\"o}rn Ommer.
\newblock High-resolution image synthesis with latent diffusion models.
\newblock In \emph{Proceedings of the IEEE/CVF conference on computer vision and pattern recognition}, pages 10684--10695, 2022.

\bibitem[Shen et~al.(2024)Shen, Yin, Wang, Wei, Cai, Yang, and Lin]{shen2024hmr}
Wenhao Shen, Wanqi Yin, Hao Wang, Chen Wei, Zhongang Cai, Lei Yang, and Guosheng Lin.
\newblock Hmr-adapter: A lightweight adapter with dual-path cross augmentation for expressive human mesh recovery.
\newblock In \emph{Proceedings of the 32nd ACM International Conference on Multimedia}, pages 6093--6102, 2024.

\bibitem[Siddiqui et~al.(2024)Siddiqui, Alliegro, Artemov, Tommasi, Sirigatti, Rosov, Dai, and Nie{\ss}ner]{siddiqui2024meshgpt}
Yawar Siddiqui, Antonio Alliegro, Alexey Artemov, Tatiana Tommasi, Daniele Sirigatti, Vladislav Rosov, Angela Dai, and Matthias Nie{\ss}ner.
\newblock Meshgpt: Generating triangle meshes with decoder-only transformers.
\newblock In \emph{Proceedings of the IEEE/CVF Conference on Computer Vision and Pattern Recognition}, pages 19615--19625, 2024.

\bibitem[Su et~al.(2024)Su, Ahmed, Lu, Pan, Bo, and Liu]{su2024roformer}
Jianlin Su, Murtadha Ahmed, Yu Lu, Shengfeng Pan, Wen Bo, and Yunfeng Liu.
\newblock Roformer: Enhanced transformer with rotary position embedding.
\newblock \emph{Neurocomputing}, 568:\penalty0 127063, 2024.

\bibitem[Tochilkin et~al.(2024)Tochilkin, Pankratz, Liu, Huang, Letts, Li, Liang, Laforte, Jampani, and Cao]{tochilkin2024triposr}
Dmitry Tochilkin, David Pankratz, Zexiang Liu, Zixuan Huang, Adam Letts, Yangguang Li, Ding Liang, Christian Laforte, Varun Jampani, and Yan-Pei Cao.
\newblock Triposr: Fast 3d object reconstruction from a single image.
\newblock \emph{arXiv preprint arXiv:2403.02151}, 2024.

\bibitem[Vaswani(2017)]{vaswani2017attention}
A Vaswani.
\newblock Attention is all you need.
\newblock \emph{Advances in Neural Information Processing Systems}, 2017.

\bibitem[Wallace et~al.(2024)Wallace, Dang, Rafailov, Zhou, Lou, Purushwalkam, Ermon, Xiong, Joty, and Naik]{wallace2024diffusion}
Bram Wallace, Meihua Dang, Rafael Rafailov, Linqi Zhou, Aaron Lou, Senthil Purushwalkam, Stefano Ermon, Caiming Xiong, Shafiq Joty, and Nikhil Naik.
\newblock Diffusion model alignment using direct preference optimization.
\newblock In \emph{Proceedings of the IEEE/CVF Conference on Computer Vision and Pattern Recognition}, pages 8228--8238, 2024.

\bibitem[Wang et~al.(2024)Wang, Zhao, Wang, Quan, and Yan]{wang2024vq}
Hanxiao Wang, Mingyang Zhao, Yiqun Wang, Weize Quan, and Dong-Ming Yan.
\newblock Vq-cad: Computer-aided design model generation with vector quantized diffusion.
\newblock \emph{Computer Aided Geometric Design}, 111:\penalty0 102327, 2024.

\bibitem[Wei et~al.(2023)Wei, Wang, Feng, Lin, and Yap]{wei2023taps3d}
Jiacheng Wei, Hao Wang, Jiashi Feng, Guosheng Lin, and Kim-Hui Yap.
\newblock Taps3d: Text-guided 3d textured shape generation from pseudo supervision.
\newblock In \emph{Proceedings of the IEEE/CVF conference on computer vision and pattern recognition}, pages 16805--16815, 2023.

\bibitem[Willis et~al.(2021{\natexlab{a}})Willis, Jayaraman, Lambourne, Chu, and Pu]{willis2021engineering}
Karl~DD Willis, Pradeep~Kumar Jayaraman, Joseph~G Lambourne, Hang Chu, and Yewen Pu.
\newblock Engineering sketch generation for computer-aided design.
\newblock In \emph{Proceedings of the IEEE/CVF conference on computer vision and pattern recognition}, pages 2105--2114, 2021{\natexlab{a}}.

\bibitem[Willis et~al.(2021{\natexlab{b}})Willis, Pu, Luo, Chu, Du, Lambourne, Solar-Lezama, and Matusik]{willis2021fusion}
Karl~DD Willis, Yewen Pu, Jieliang Luo, Hang Chu, Tao Du, Joseph~G Lambourne, Armando Solar-Lezama, and Wojciech Matusik.
\newblock Fusion 360 gallery: A dataset and environment for programmatic cad construction from human design sequences.
\newblock \emph{ACM Transactions on Graphics (TOG)}, 40\penalty0 (4):\penalty0 1--24, 2021{\natexlab{b}}.

\bibitem[Wu et~al.(2021)Wu, Xiao, and Zheng]{wu2021deepcad}
Rundi Wu, Chang Xiao, and Changxi Zheng.
\newblock Deepcad: A deep generative network for computer-aided design models.
\newblock In \emph{Proceedings of the IEEE/CVF International Conference on Computer Vision}, pages 6772--6782, 2021.

\bibitem[Xu et~al.(2021)Xu, Peng, Cheng, Willis, and Ritchie]{xu2021inferring}
Xianghao Xu, Wenzhe Peng, Chin-Yi Cheng, Karl~DD Willis, and Daniel Ritchie.
\newblock Inferring cad modeling sequences using zone graphs.
\newblock In \emph{Proceedings of the IEEE/CVF conference on computer vision and pattern recognition}, pages 6062--6070, 2021.

\bibitem[Xu et~al.(2022)Xu, Willis, Lambourne, Cheng, Jayaraman, and Furukawa]{xu2022skexgen}
Xiang Xu, Karl~DD Willis, Joseph~G Lambourne, Chin-Yi Cheng, Pradeep~Kumar Jayaraman, and Yasutaka Furukawa.
\newblock Skexgen: Autoregressive generation of cad construction sequences with disentangled codebooks.
\newblock \emph{arXiv preprint arXiv:2207.04632}, 2022.

\bibitem[Xu et~al.(2023)Xu, Jayaraman, Lambourne, Willis, and Furukawa]{xu2023hierarchical}
Xiang Xu, Pradeep~Kumar Jayaraman, Joseph~G Lambourne, Karl~DD Willis, and Yasutaka Furukawa.
\newblock Hierarchical neural coding for controllable cad model generation.
\newblock \emph{arXiv preprint arXiv:2307.00149}, 2023.

\bibitem[Xu et~al.(2024)Xu, Lambourne, Jayaraman, Wang, Willis, and Furukawa]{DBLP:journals/tog/XuLJWWF24}
Xiang Xu, Joseph~G. Lambourne, Pradeep~Kumar Jayaraman, Zhengqing Wang, Karl D.~D. Willis, and Yasutaka Furukawa.
\newblock Brepgen: {A} b-rep generative diffusion model with structured latent geometry.
\newblock \emph{{ACM} Trans. Graph.}, 43\penalty0 (4):\penalty0 119:1--119:14, 2024.

\bibitem[Yang et~al.(2019)Yang, Huang, Hao, Liu, Belongie, and Hariharan]{yang2019pointflow}
Guandao Yang, Xun Huang, Zekun Hao, Ming-Yu Liu, Serge Belongie, and Bharath Hariharan.
\newblock Pointflow: 3d point cloud generation with continuous normalizing flows.
\newblock In \emph{Proceedings of the IEEE/CVF international conference on computer vision}, pages 4541--4550, 2019.

\bibitem[Ye et~al.(2023)Ye, Zhang, Liu, Han, and Yang]{ye2023ip}
Hu Ye, Jun Zhang, Sibo Liu, Xiao Han, and Wei Yang.
\newblock Ip-adapter: Text compatible image prompt adapter for text-to-image diffusion models.
\newblock \emph{arXiv preprint arXiv:2308.06721}, 2023.

\bibitem[Yin et~al.(2023)Yin, Zhang, Chen, Cai, Yu, Wang, Chen, and Shen]{yin2023metric3d}
Wei Yin, Chi Zhang, Hao Chen, Zhipeng Cai, Gang Yu, Kaixuan Wang, Xiaozhi Chen, and Chunhua Shen.
\newblock Metric3d: Towards zero-shot metric 3d prediction from a single image.
\newblock In \emph{Proceedings of the IEEE/CVF International Conference on Computer Vision}, pages 9043--9053, 2023.

\end{thebibliography}
}

  \clearpage
\setcounter{page}{1}
\maketitlesupplementary

\section{Supplementary Materials}
\label{sec: supplementary}
We have prepared supplementary materials. The technical details of our implementation are discussed in~\cref{sec:tech_details} and ~\cref{sec:data_details}. Moreover, we present additional examples and comparisons in~\cref{sec:more_res} to demonstrate the performance of our method.

\section{Technical Details}
\label{sec:tech_details}

\subsection{CAD Commands Encoding} We define the CAD command sequence following DeepCAD \cite{wu2021deepcad}, focusing on the two commonly used categories: \emph{sketch} and \emph{extrusion}, where \emph{sketch} includes commands $\mathtt{start} \{\cmdSOL\}$, $\mathtt{line} \{ \cmdLine\}$, $\mathtt{arc} \{\cmdArc\}$, and $\mathtt{circle} \{\cmdCirc\}$ and \emph{extrusion} has a single command $\cmdExt$, we also need an end command $\cmdEOS$ for the entire command sequence.
Each command is defined by a few parameters for their location, size, and orientation. The detailed definitions of the parameters are given in Table \ref{tab:data_rep}. For the $i$-th line of command $C_i = (s_i, p_i)$, where $s_i$ is the command type and we stack all the parameters for all command types into a vector $p_i = [x, y, \alpha, f, r,\theta, \phi, \gamma, p_x, p_y, p_z, s, e_1, e_2, b, u]$, setting unused parameters to $-1$. We then pad the sequence to a fixed length, $\cmdNum=60$, using the empty command $\cmdEOS$.

\begin{table}[h]
\centering
\resizebox{\columnwidth}{!}{%
\begin{tabular}{ccc}
\toprule
 \textbf{Commands} & \textbf{Parameters} \\ \toprule
$\cmdSOL$ & $\emptyset$ \\ \hline
 \makecell{$\cmdLine$ \\ (Line)} & \makecell[l]{
  $\begin{aligned} 
 \quad \quad\; x, y: \text{end-points of line}
 \end{aligned}$
 }\\ \hline
 \makecell{$\cmdArc$ \\ (Arc)} & \makecell[l]{
 $\begin{aligned}
 \quad \quad\; x, y &: \text{end-points of arc} \\ \alpha &: \text{sweep angle} \\ f &: \text{flag for counter-clockwise}
 \end{aligned}$
 }\\ \hline
 \makecell{$\cmdCirc$ \\ (Circle)} & \makecell[l]{
 $\begin{aligned}
 \quad \quad\; x, y &: \text{center of circle}\\ r &: \text{radius of circle}
 \end{aligned}$
 }\\ \hline
 \makecell{$\cmdExt$ \\ (Extrude)} &  \makecell[c]{
 $\begin{aligned}
 \theta, \phi, \gamma &: \text{orientation of sketch plane}\\ p_x, p_y, p_z &: \text{origin of sketch plane} \\ s &: \text{associated sketch profile scale} \\ e_1, e_2 &: \text{extrude distances toward both sides}\\ b &: \text{bool type},\quad u : \text{extrusion type}
 \end{aligned}$
 }\\ \hline
$\cmdEOS$ & $\emptyset$  \\ \bottomrule
\end{tabular}}
\caption{The CAD commands and parameters defined in DeepCAD \cite{wu2021deepcad} convention.} 
\label{tab:data_rep}

\end{table}

\subsection{CAD Autoencoder} 
 Our autoencoder architecture is similar to  \cite{wu2021deepcad}. We formulate the task as a classification problem to simplify the learning process. We normalize all CAD models and quantize the continuous parameters into 256 levels represented as 8-bit integers. Therefore, each parameter $p_{i,j}$ where $j \in \{1\cdots16\}$ is represented by a one-hot embedding of dimension $256+1=257$ with an additional element reserved for unused parameters. We tokenize the commands by mapping them to embedding spaces with learnable matrices, the resulting embedding $e(C_i) = \embCmd+\embPm+\embPos\in\mathbb{R}^{\embDim}$, where $\embPos$ is a learnable positional embedding and $\embDim=256$ is the embedding dimension. The embedding is passed through four layers of transformer blocks and we take the averaged outputs as the latent vector $z$ with the same dimension 
 $\embDim=256$. Then, we reconstruct the CAD command sequence from the latent vector $z$ through a decoder with the same structure as the encoder followed by two linear prediction heads for commands $s_i$ and parameters $p_i$. The training objective of the autoencoder is to learn accurate predictions of CAD parameters and to regularize the latent space. The training loss is defined as a cross-entropy loss between the predicted $\hat{C}$ and ground-truth $C$.

\subsection{Discussion on Regularization of Autoencoder.}\label{Regularization_of_Autoencoder}  In addition to the reconstruction loss mentioned above, to further regularize the generated latent space, we have also experimented with different regularization terms. For example, we use the KL divergence as a regularization term: $l_{\text{kl}} = D_{\text{KL}}(q(z|C_i) \parallel p(z))$. In this equation, \( D_{\text{KL}} \) represents the Kullback-Leibler divergence, \( q(z|C_i) \) is the latent distribution conditioned on the input \( C_i \), and \( p(z) \) is the prior distribution of the latent space. This regularization term ensures that the encoded latent representation closely approximates the predefined prior distribution, which is set as a Gaussian distribution with zero mean and a standard deviation of 0.25. We also utilize a constant \(\beta\) to adjust the strength of the regularization, setting its value to \(1 \times 10^{-5}\). The VAE reconstruction results are shown in Table \ref{table_ae}, demonstrating that the model can reconstruct the sequence with high precision in both scenarios. The regularization terms have minimal impact on the results. Moreover, using the regularization term to train the diffusion model does not result in improvements, so our AE is only trained using the reconstruction loss. To obtain representations better suited for latent diffusion, future work could potentially increase the latent capacity, such as using a sequence of latent instead of a single latent.

\begin{table}\setlength{\tabcolsep}{1.0pt}
\fontsize{9}{10}\selectfont
\centering
\begin{tabular}{lccccccc}
\toprule
        \textbf{Methods} & $\text{ACC}_{\text{cmd}} \uparrow$ & $\text{ACC}_{\text{para}} \uparrow$ & \textbf{Med CD} $\downarrow$ & \textbf{IR} $\downarrow$ \\ 
        \midrule
        AE$_{w/o-L_{kl}}$  & 99.52 & 98.18 & 0.073 & 0.026 \\
        AE$_{w-L_{kl}}$  & 99.32 & 98.02 & 0.075 &  0.027 \\

\midrule
\end{tabular}
\vspace{-10pt}
\caption{Quantitative evaluation of different autoencoding strategies. The CD is multiplied by
$10^2$.}\label{table_ae}
\vspace{-15pt}
\label{components}
\end{table}

\subsection{Diffusion Transformer Network} 
Our diffusion transformer architecture follows DALLE-2 \cite{ramesh2022hierarchical}, comprising 12 blocks, each containing a self-attention layer and a fully connected layer. During testing, we start with a randomly sampled noise vector \( z_T \) drawn from a standard normal distribution \( \mathcal{N}(0, I) \). Our diffusion model is then iteratively applied to this vector to progressively denoise it, resulting in the final output \( z_0 \). This process is described by:
\begin{equation}
\small z_0 = (f \circ \dots \circ f)(z_T, T, f_m), \quad f(x_t, t) = \Omega(x_t, \gamma(t)| f_m) + \sigma_t \epsilon,
\end{equation}

where \( \sigma_t \) represents the fixed standard deviation at each timestep \( t \), and \( \epsilon \) is sampled from \( \mathcal{N}(0, I) \). We continue to denoise \( z_T \) through successive iterations until \( z_0 \) is achieved. The resulting latent vectors \( z_0 \) are then fed into the previously trained decoder to reconstruct the CAD sequence. We employ the DDPM solver \cite{ho2020denoising}. Since our training objective function is to predict \( x_0 \), we can rearrange the equation of the forward diffusion process to compute \( \epsilon \) from \( x_0 \). This allows us to predict the noise \( \epsilon \) directly based on the predicted \( x_0 \).
\begin{figure}[t]
  \centering
   \includegraphics[width=\linewidth]{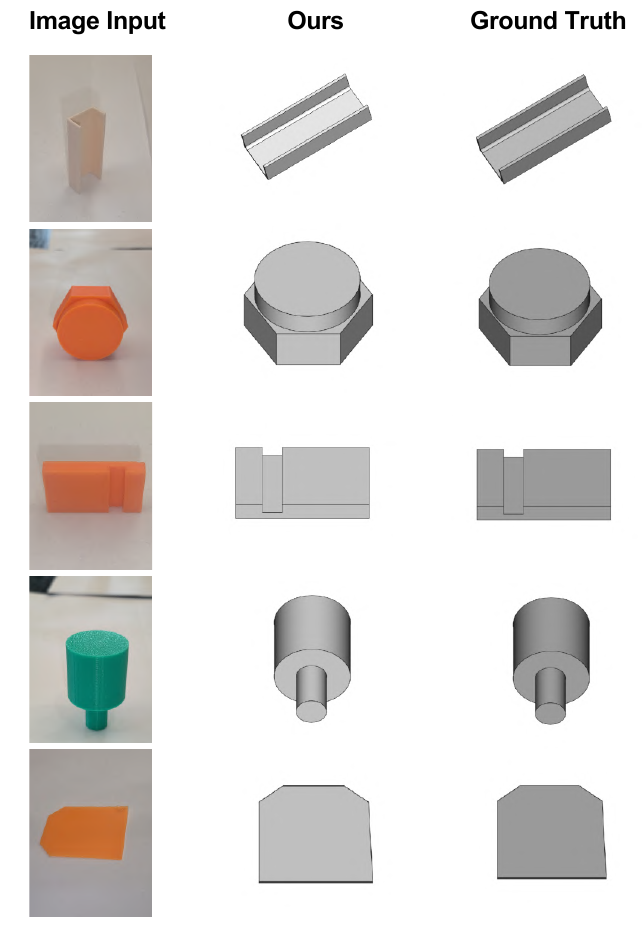}
\vspace{-10pt}
   \caption{More generated results on RealCAD dataset by our method, the real images are shown on the left.}
   \label{fig:supp_real}
   \vspace{-10pt}
\end{figure}
\section{Dataset Details}
\label{sec:data_details}
We render the compiled CAD models using Blender. To provide comprehensive multi-view information while accommodating our unconstrained testing scenario, for each model, we generate eight sets of four-view images. In each set, we sample four camera locations with mean azimuth angles separated by 90 degrees, applying a random perturbation within a 30-degree range to each azimuth. The four views share the same randomly chosen elevation angle and a radius sampled from 1.8 to 2.5 units. Additionally, for each set, the CAD object is randomly rotated within a range of -15 to 15 degrees along each axis.

While collecting our RealCAD dataset, the collector casually captured images of the object from approximately four different angles: front-left, front-right, back-left, and back-right. There were no specific requirements regarding the elevation and radius for these shots. The 3D-printed CAD models, featuring a variety of textures and colors, were photographed under standard indoor lighting conditions using iPhones.

\section{More Results }

\begin{figure}[t]
  \centering
   \includegraphics[width=\linewidth]{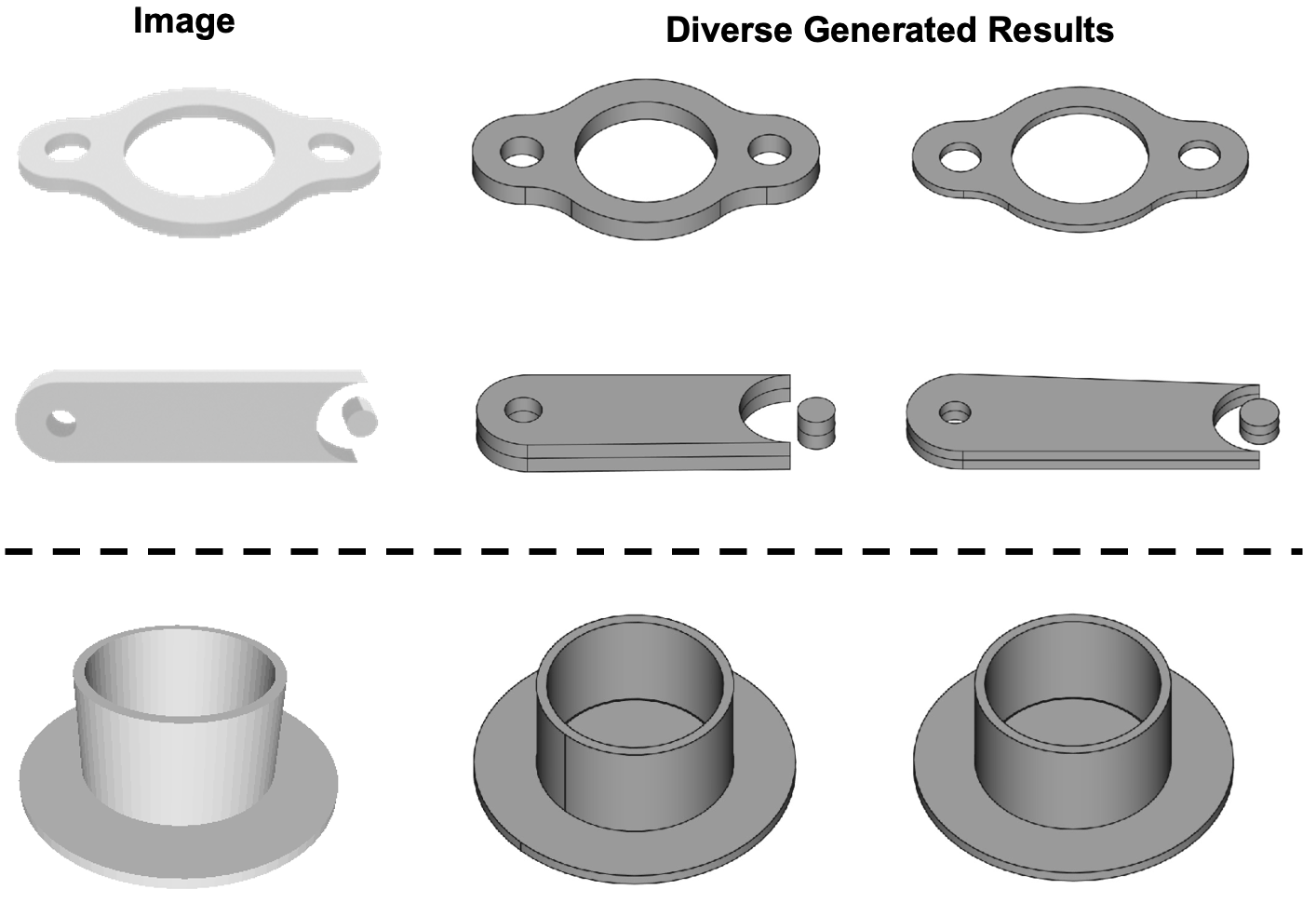}
\vspace{-10pt}
   \caption{Diverse generated results with multi-view input. To simplify, we use a single image to represent multi-view inputs. Our model reliably captures geometric details, with occasional size variations (upper part). It also generates diverse designs, such as representing a circle as either a full circle or two semi-circular curves (lower part).
}
   \label{fig:supp_multi-view-diverse}
   \vspace{-10pt}
\end{figure}
\label{sec:more_res}
\subsection{Multi-View Reconstruction Diversity}
In Figure \ref{fig:single_multi_view} of the main text, we showcase the diverse results generated using a single view as input. In the single-view setting, our model can produce results with varying levels of complexity for the unseen parts of the object. This is because, with only one view, the model infers the hidden regions, leading to diversity in the generated outputs.

When we switch to the multi-view setting, the multiple perspectives provide comprehensive information about the object. Consequently, the generated results typically present a complete reconstruction of the object's shape, differing mainly in size. As shown in the upper part of Figure \ref{fig:supp_multi-view-diverse}, we provide examples generated using multi-view inputs. Across different sampling runs, our model consistently recovers the object's shape. However, due to the inherent ambiguity in the image data regarding object scale, the generated results exhibit variations in size. Additionally, our method can generate various CAD design sequences for the same model. As shown in the lower part of Figure \ref{fig:supp_multi-view-diverse}, the generated circle may be represented as either a full circle or two semi-circular curves. 
\begin{table}\setlength{\tabcolsep}{1.0pt}
\fontsize{9}{10}\selectfont
\centering
\begin{tabular}{lccccccc}
\toprule
        \textbf{Methods} & $\text{ACC}_{\text{cmd}} \uparrow$ & $\text{ACC}_{\text{para}} \uparrow$ & \textbf{Med CD} $\downarrow$ & \textbf{IR} $\downarrow$ \\ 
        \midrule
        CADCrafter$_{zero123}$  & 63.89 & 42.98 & 0.201 & 0.466 \\
        CADCrafter  & 84.62 & 73.31 & 0.026 & 0.036 \\

\midrule
\end{tabular}
\vspace{-10pt}
\caption{ Performance comparisons of the multi-view diffusion model on the DeepCAD dataset.}\label{table_zero123_comare}
\vspace{-15pt}
\label{components}
\end{table}
 
\subsection{Discussion on Multi-View Diffusion}
In our architecture, we employ a distillation loss to enable our single-view geometry encoder to learn from multi-view knowledge. We have also explored an alternative approach where a multi-view diffusion model is directly employed to generate images from different views using a single-view input. For this experiment, we fine-tune the Zero-1-to-3 model \cite{liu2023zero} using our rendered CAD image dataset. Despite this effort, the multi-view diffusion model struggled to accurately capture geometry across different views, introducing noise during the conditioning process and ultimately degrading overall performance. We quantitatively evaluate this method on DeepCAD, and the results shown in Table \ref{table_zero123_comare} further underscore the necessity of our designs.

\subsection{More Results on RealCAD} Here,  we showcase more generated results on the RealCAD dataset by our method in Figure \ref{fig:supp_real}. It can be observed that our model handles different object poses and sizes effectively. For instance, in the last row, even for very thin objects, the parameters are generated correctly.

\end{document}